# Residual-CNDS for Grand Challenge Scene Dataset

Hussein A. Al-Barazanchi, Hussam Qassim, David Feinzimer, *and* Abhishek Verma
Department of Computer Science
California State University
Fullerton, California 92834
Email: {hussein_albarazanchi, hualkassam, dfeinzimer}(at)csu.fullerton.edu, averma(at)fullerton.edu

*Abstract*— *Increasing depth of convolutional neural networks (CNNs) is a highly promising method of increasing the accuracy of the (CNNs). Increased CNN depth will also result in increased layer count (parameters), leading to a slow backpropagation convergence prone to overfitting. We trained our model (Residual-CNDS) to classify very large-scale scene datasets MIT Places 205, and MIT Places 365-Standard. The outcome result from the two datasets proved our proposed model (Residual-CNDS) effectively handled the slow convergence, overfitting, and degradation. CNNs that include deep supervision (CNDS) add supplementary branches to the deep convolutional neural network in specified layers by calculating vanishing, effectively addressing delayed convergence and overfitting. Nevertheless, (CNDS) doesn't resolve degradation; hence, we add residual learning to the (CNDS) in certain layers after studying the best place in which to add it. With this approach we overcome degradation in the very deep network. We have built two models (Residual-CNDS 8), and (Residual-CNDS 10). Moreover, we tested our models on two large-scale datasets, and we compared our results with other recently introduced cutting-edge networks in the domain of top-1 and top-5 classification accuracy. As a result, both of models have shown good improvement, which supports the assertion that the addition of residual connections enhances network CNDS accuracy without adding any computation complexity.*

## I. INTRODUCTION

Lately, convolutional neural networks have made great advancement in computer vision competitions (ILSVRC) [1, 2], which is ImageNet Large Scale Visual Recognition Challenge, one of the best proving grounds for the computer vision algorithms. Also, (CNNs) have shown a great breakthrough in other segments of image classification jobs [3, 4]. The convolutional neural networks have the power to evaluate different features of the images, and classify them in a comprehensive framework. The accuracy of the features that is extracted can be elevated by increasing the layer count being used to build the (CNN). The CNN accuracy could be enhanced by adding more layers to the network (going deep), a method proven in the ILSVRC competition [6, 7]. Furthermore, the highest scores achieved in [6-9] were all using quite deep CNN models on the ImageNet dataset [10]. Hence, network depth is of great significance. The effeteness of very deep convolutional neural can be seen in image classification and even more complicated tasks like detecting objects and performing segmentation [11-15]. However, very deep models with increased layer count raise certain problems such as slow convergence, overfitting, and vanishing/exploding gradients [16, 17].

There were several approaches in trying to resolve the problems of slower convergence and overfitting. One of these, first introduced by Simonyan and Zisserman [6], utilizes the pre-trained weights derived from networks of less depth to set the weights of very deep networks. However, using this technique has shown to be computationally expensive, and the difficulty in tuning the parameters. Szegedy et al. [7] also proposed another technique to solve the same problems: delayed convergence and overfitting. Where their proposed technique used auxiliary branches connected to shallower layers of the neural network. The objective of Szegedy et al. [7] had in utilizing these auxiliaries was to have them work as classifiers to boost the gradients for reversed propagation over layers of the deep neural network. Moreover, the auxiliary branches encourage feature maps in the middle layers. Nevertheless, Szegedy et al. [7] did not have a specified process defined the location of where or how to add the auxiliary branches. A similar approach was proposed by Lee et al. [18] in which they add auxiliary branches next to every shallower layer, which the losses propagate through these auxiliary branches summing with the loss of the output layer. Lee et al. [18] have shared some good results and enhanced the rate of the convergence, but Lee et al. [18] did not utilize the idea of deeply supervised networks (DSN) [18] with deep architecture.

To address the problems of slower convergence and overfitting, Wang et al. [19] proposed the idea of convolutional

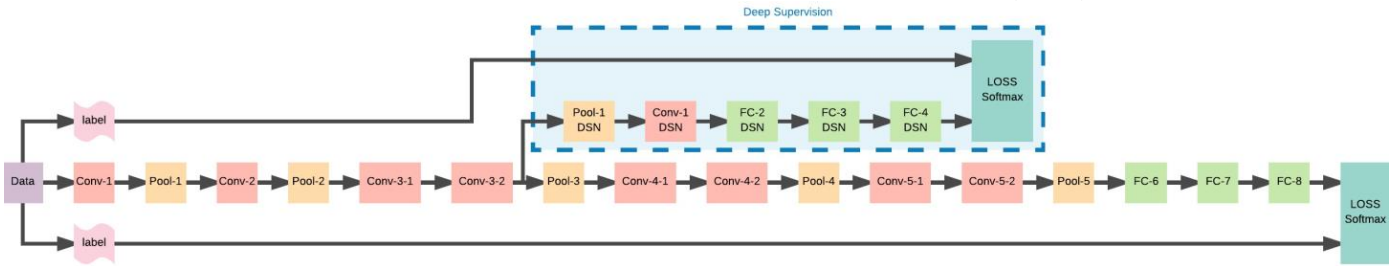

FIG (1)

THE STRUCTURE OF EIGHT CONVOLUTIONAL NETWORKS WITH DEEP SUPERVISION (CNDS) [19]

neural networks with deep supervision (CNDS), whom handle the task of determining where to attach supplementary branches. Wang et al. [19] studied vanishing gradients in the deep network, specifically having them decide where to add the branches. Determining auxiliary branch locations has solved the problem of overfitting and slower convergence. However, the degradation problem in very deep networks still went without resolution. The degradation problem arises whenever the network goes deeper, saturating the accuracy of the deep neural networks leading to a rapid collapse. The degradation problem makes the neural networks prone to high training error as reported in [20, 21] as more and more layers are added to the network; furthermore, overfitting is not the cause of the degradation issue. As the result of studying degradation, the researchers have shown that various networks are not all easily optimized. In this paper, we target slower convergence, degradation and overfitting at once through building a deep network integrated on the CNDS network structure with residual learning [22], a state of the art technique in handling degradation. We built two networks, each of which has residual connections [22] integrated with the CNDS [19] structure. We trained our two models on the very large-scale MIT Places 205 and Places365-Standard scene datasets. We also compared our networks with AlexNet [2] and GoogLeNet [7] on both MIT Places 205 and Places365-Standard scene datasets. AlexNet was proposed by Krizhevsky et al. [2] in 2012. Krizhevsky et al. [2] performed training on a deep neural network consisting of five conv. layers and three fully-connected layers. AlexNet [2] achieved a good result in the ImageNet LSVRC-2010 contest, a result which outperformed previous state-of-the-art networks. Consistent with AlexNet [2], GoogLeNet [7] was a state-of-the-art network when it was first proposed by Szegedy et al. [7]. GoogLeNet [7] achieved a high result in the ImageNet Large-Scale Visual Recognition Challenge 2014 (ILSVRC14) in classification and detection. Szegedy et al. [7] improved computational resource utilization methods inside the network of their architecture, in which they increased the width and the depth of the neural network; however, computational complexity remained unchanged. Our proposed models have shown a better result in classification than both AlexNet [2] and GoogLeNet [7] in top1 and top5 on both MIT Places 205 and Places365-Standard. However, our models are/almost are less computationally complex than GoogLeNet [7] and AlexNet [2]. Furthermore, the two networks eight and ten convolutional layers have shown the advantage of integrating residual learning [22] and CNDS [19], which increases the accuracy of CNDS network.

Section II will provide background on the CNDS network and residual learning. Section III will provide the specifications of the prospective Residual-CNDS. Section IV presents the details of the MIT Places 205 and Places365-Standard scene oriented datasets on which our experiments were ran. Section V describes our procedural methods. We further discuss our results in section VI.

## II. BACKGROUND

At ILSVRC 2014 contest [1], problems concerning computer vision were used to benchmark networks, and it is where the notion behind utilizing very deep artificial neural networks first acquired importance. Therefore, the study of efficient ways to train very deep neural networks has as of recent demonstrated indications of progress. In part (A) we discuss the structure of CNDS network and the way it's authors use vanishing gradients to determine the fit location to insert the supplementary branch. Following that, we explore the residual learning mechanism. The exploration and discussion of CNDS and residual learning techniques provides a fundamental background and supplies a useful basis for the proposed networks (Residual-CNDS) that are going to be explained at a later point.

### A. CNDS Network

Szegedy et al. [7] introduced the idea of adding subsidiary classifiers whom link to the middle layers. The neural network's generalization gets much better by adding these auxiliary classifiers, whom provide more supervision in the training stage. Nevertheless, Szegedy et al. [7] doesn't provide documentation on where to add the subsidiary classifier. Lee et al. [18] discussed where to add the classifier branches. The Lee et al. [18] network, called deeply-supervised nets (DSN), connects am SVM classifier to the output of each hidden layer of the network. Lee et al. [18] utilize this mechanism in the training mode. They achieve optimization through a summation of loss from the output layer and that of subsidiary classifiers.

The matter of where to add the auxiliary classifiers has been handled by Wang et al. [19]. Wang et al. [19] deep supervision

FIG (2)

PROPOSED EIGHT CONVOLUTIONAL NETWORKS WITH DEEP SUPERVISION AND RESIDUAL LEARNING (RESIDUAL-CNDS). DASHED RECTANGLE SHOWS THE DEEP SUPERVISION BRANCH AND
RESIDUAL CONNECTIONS ARE THE RED

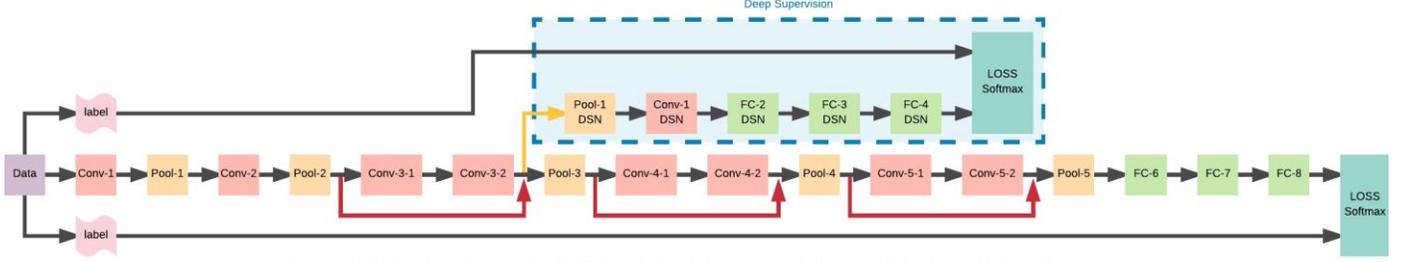

(CNDS) networks have some big distinctions from that of Lee et al. [18]. In the beginning, Lee et al. [18] connects the branch classifier in every instance of a hidden layer, while Wang et al. [19] utilizes a gradient-based heuristic method in determining whether or not an auxiliary classifier. Another variation in the Wang et al. [19] paper, is that they utilize a tiny artificial neural network as a subsidiary supervision classifier. This small branch contains convolutional layer, some fully connected layers, all followed by a Softmax, a design which closely resembles Szegedy et al. [7]. On the other hand, SVM classifiers, which are linked to hidden layer output were used by Lee et al. [16].

The vanishing of gradients process was used by Wang et al. [19] in determining auxiliary supervision branch locations. In the beginning, Wang et al. [19] built the neural network without the supervision classifiers. Weights of the neural network were set from Gaussian pattern with a zero mean, standard deviation of 0.01, and bias of 0. Wang et al. [19] then perform between 10 and 50 back-propagation epochs and control the mean gradient amount of the shallower layers by plotting the subsidiary supervision classifiers add whenever the mean gradient rate degrades, i.e., drops under a $10^{-7}$ threshold. In the Wang et al. [19] approach, the average gradient drops under the appointed threshold. Appropriately, the auxiliary classifier was added following the fourth layer by Wang et al. [19]. Fig (1) demonstrates the Wang et al. [19] network approach. In this paper, to make an easier comparison between our proposed design and Wang et al. [19] design, we follow the Wang et al. [19] paper's naming style.

### Residual Learning

Degradation decays the optimization in deep convolutional neural networks. Increasing depth of the convolutional neural networks should increase the networks accuracy. Nevertheless, the empirical result demonstrates that the error presented by the deeper convolutional neural networks is higher than their equivalent superficial neural networks. The degradation problem has been solved by the proposed He et al. [22] design. The residual learning proposed by He et al. [22] let every few stacked layer qualify as a residual mapping as degradation stops layers from fitting the required subsidiary mapping. They alter the subsidiary mapping to resemble that of formula (2) rather than formula (1). He et al. [22] assumes a harder optimization of the primary mapping than the residual mapping.

$$F(x) = H(x) \qquad (1)$$

$$F(x) = H(x) - x \qquad (2)$$

$$F(x) = H(x) + x \qquad (3)$$

Shortcut links are expressed by formula 3 [22]. A shortcut connection serves as a process of surpassing (x >= 1) layers in the convolutional neural network [23-25]. Fig (3) demonstrates the way that the shortcut connection can be performed in a convolutional neural network. He et al. [22] utilize the concept of shortcut connections to perform identity mapping [22] as in Fig (3). The shortcut connections output is fused with stacked layer output as shown in Fig (3), as shown in formula (3). Identity shortcut connections have the advantage of being parameter free. Highway networks [21] have shown a difference, as shortcut connections are presented with parameterized gating functions [26]. Another advantage of the He et al. [22] is that through the stochastic gradient descent algorithm we can optimized identity shortcut connections as an end to end solution. Furthermore, the identity shortcut connections are easy to execute when utilizing deep learning libraries like [27-30].

### III. PROPOSED RESIDUAL-CNDS NETWORK ARCHITECTURE

In this work, the first proposed Residual-CNDS version's main branch contains eight conv. layers. We utilize filters with a micro receptive area, only 3*3, the smallest size possible when trying to grasp the idea of (left/right, up/down, center). Convolution layer stride is fixed to 1 pixel in the main branch except for the first layer which has a stride of two with a convolutional layer padding of 1 pixel in the main branch.

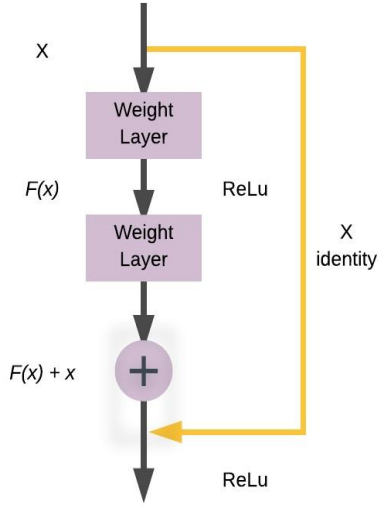

FIG (3)

RESIDUAL CONNECTION [22]

Furthermore, we added a Scale layer to each convolutional neural layer in the main branch and spatial pooling is performed by five max-pooling layers, whom accompany some conv. layers. It should be noted that not every convolutional layer is accompanied by max-pooling layers. All max-pooling layers are implemented on a 2*2-pixel window with the exception of the first which is done on a 3*3-pixel window, with a stride of 2. Applying rules from Wang et al. [19], we attach the auxiliary classifier following the convolutional layer which is prone to the vanishing gradients problem (conv-3-2), as shown in Fig (1). Feature maps, created in the shallower layers as Wang et al. [19] describes, are noisy, requiring us to minimize the noise production in the shallower layers prior to passing them to the classifiers. For that reason, we initialize our subsidiary branch with an average pooling layer that has 5*5 kernel size and a stride of two. Following this, we add a convolutional layer with a kernel size of 1*1 and a stride of one. Finally, we attach two fully connected layers, both containing 1,024 channels and dropout ratio of 1/2. In comparison with the subsidiary branch, the main branch contains two F.C. layers with 4,096 channels as well as a dropout ratio of ½ for each fully connected layer. All hidden and fully connected layers are supplied with the rectification (ReLU (Krizhevsky et al., 2012)) non-linearity, with the exception of the convolutional layer preceding the element-wise addition connections, the branch convolutional layer and the fully connected output layers. Lastly, the main and the subsidiary branch each have a dedicated output layer, containing a fully connected layer succeeded by a softmax layer to determine probability of a class.

$$W_{main} = (W_1, \ldots, W_{11}) \quad (4)$$
$$W_{branch} = (W_{s5}, \ldots, W_{s8}) \quad (5)$$

Equation (4) [19] shows the names of main branch weights. Moreover, the equation's weights (4) resemble three fully connected layers and eight convolutional layers. Equation (5) [19] illustrates the names of weights of the subsidiary branch whom resemble the three fully connected layers and the first convolutional layer. If $X_{11}$ is assigned to symbolize the map of features derived from the output layer in the main branch, then we can compute probability from the softmax function from the labels k =1, ..., K, a formula depicted by equation (6) [19]. Furthermore, if we assign $S_8$ to represent the feature map that is derived from the output layer in the auxiliary classifier, then calculating return can be done with the help of equation (7) [19].

$$pk = \frac{\exp(X_{11(k)})}{\sum_k \exp(X_{11(k)})} \quad (6)$$

$$psk = \frac{\exp(S_{8(k)})}{\sum_k \exp(S_{8(k)})} \quad (7)$$

The loss, computed from the main branch is illustrated in equation (8) [19], which calculates likelihoods generated by the softmax function. Also, loss generated from the subsidiary classifier is computed by using equation (9) [19]. Furthermore, the loss generated from the subsidiary classifier carries the weights of the subsidiary classifier and the weights of the previous convolutional layers before the subsidiary classifier in the main branch.

$$L_0(W_{main}) = -\sum_{k=1}^{K} yk \ln pk \quad (8)$$
$$L_s(W_{main}, W_{branch}) = -\sum_{k=1}^{K} yk \ln psk \quad (9)$$

Loss generated from the main branch and the subsidiary classifier can be computed using equation (10) [19]. Equation 10 computes weighted total as the main branch is illustrated with greater weight in comparison to the auxiliary classifier. Furthermore, $\alpha_t$ represents the rate of the auxiliary classifier as an adjustment factor. Symbol $\alpha_t$ represents decay with multiple iterations as shown in equation (11) [19], with N representing total iteration count.

$$L_s(W_{main}, W_{branch}) = L_0(W_{main}) + \alpha_t L_s(W_{main}, W_{branch}) \quad (10)$$

$$\alpha_t = \alpha_t * (1 - t/N) \quad (11)$$

The residual learning design we present adapts the shortcut connections proposed by He et al. [22] depicted below by equation (12).

TABLE (1)

COMPARISON OF THE TOP 1 & 5 VALIDATION AND TEST CLASSIFICATION ACCURACY (%) WITH OTHER PEER REVIEWED PUBLISHED NETWORKS ON THE MIT PLACES 205 DATASET

| Network | Top-1 Validation/Test | Top-5 Validation/Test |
|---|---|---|
| AlexNet [2] | - / 50.0 | - / 81.1 |
| GoogLeNet [7] | - / 55.5 | - / 85.7 |
| CNDS-8 [19] | 54.7 / 55.7 | 84.1 / 85.8 |
| Our Model: Residual-CNDS-8 | 55.61/ 57.03 | 84.78/ 86.46 |

Note: Data not available is marked as '- '

$$y = F(x, \{W_i\}) + x \quad (12)$$

Many experiments have concentrate on the CNDS architecture, the process of deciding where to add residual connections [22] and the process of deciding how many layers should be skipped in shortcut connections [22]. We made a decision to add residual connections [22] to the main branch exclusively. We cannot add residual connections to the auxiliary classifier because there are no consecutive convolutional layers in the auxiliary classifier. The proposed design, shown in Figure 2, has demonstrated residual learning connections in the master branch. From here on out we will symbolize convolutional layers with conv., pooling layers with Pool., and fully connected layers with FC. In the beginning, the residual connection fuses input of the Conv3-1 to the output of Conv3-2 using element-wise addition connection, as element-wise links the output of Pool-2 to the output of Conv3-2. Conv-2 and Conv3-2 have a kernel size of 128 and 256 respectively. To give Conv2 and Conv3_2 a kernel of equal size we add a convolutional layer with a kernel of size 256 between Pool-2 and element-wise addition. Next, we add the second residual connection following Pool-3 while the shortcut connection surpasses conv. layers Conv4_1 and Conv4_2. Hence, the residual connection is linked to the output of Pool-3 and the output of Conv4-2. The kernel of Conv3-2 is of size 256 and Conv4-2 has a kernel of size 512, calling for us to a convolutional layer with kernel of size 512. The convolutional layer added between Pool-3 and the element-wise addition layer regulates the size of Pool-3 and Conv4-2 kernels, as demonstrated in Figure 4. Furthermore, the auxiliary classifier is attached after the element-wise addition between the output of Pool-3 and Conv4-2. The final residual connection connects output of Pool-4 and the latest convolutional layer, Conv5-2. In the third residual connection, it is not necessary to add a modification convolutional layer between Pool-4 and element-wise addition as the kernel size of Conv4-2 and Conv5-2 is 512.

The second proposed Residual-CNDS version two contains a master branch with 10 conv. layers. After following the same process we described for Residual-CNDS version one, we add a Scale layer to each convolutional layer in the master branch. The first convolution layer has assigned kernel size 7*7 and stride of two. The rest of the convolutional layers in the Residual-CNDS version two have a kernel size of 3*3 with a stride of one in the master branch. Following the rule proposed by Wang et al. [19], the auxiliary classifier, which has the supervision branch, is attached after the convolutional layer (Conv-3-2), the same convolutional layer suffering from vanishing gradients issue, as demonstrated in Figure 5. The feature maps generated by the shallower layers appear to be noisy. It is very important to reduce the noise in these convolutional layers before the classifiers are reached. We minimize feature map dimensionality [19] and we pass them into non-linear functions prior to placing them into classifiers. Accordingly, the auxiliary branch begins with an average pooling layer that has kernel of size five and stride of two followed by a convolutional layer with a kernel size of 1*1 and stride of one. Later, we add two fully connected layers with 1,024 output channels followed with a ½ ratio dropout layer for each fully connected layer. The second subsidiary branch in the Residual-CNDS version two is added following convolutional layer (Conv-3-2). The second subsidiary branch has the same architecture and values of the first subsidiary classifier. The main branch has two fully connected layers of channel size of 4,096 attached to both of them and ½ ration dropout layers. All hidden and fully connected layers are supplied with the rectification (ReLU (Krizhevsky et al., 2012)) non-linearity, with the exception of the convolutional layer prior to the element-wise addition

TABLE (2)

COMPARISON OF THE TOP 1 & 5 VALIDATION CLASSIFICATION ACCURACY (%) WITH OTHER PEER REVIEWED PUBLISHED NETWORKS ON THE MIT PLACES365-STANDARD DATASET

| Network | Top-1 Validation | Top-5 Validation |
|---|---|---|
| AlexNet [2] | 47.51 | 78.03 |
| GoogLeNet [7] | 50.88 | 81.49 |
| CNDS-8 [19] | 50.68 | 81.17 |
| Our Model: Residual-CNDS-8 | 51.93 | 82.25 |
| Our Model: Residual-CNDS-10 | 51.92 | 82.42 |

FIG (4)

SPECIFICS OF THE RESIDUAL CONNECTIONS IN THE PROPOSED (RESIDUAL-CNDS) NETWORK

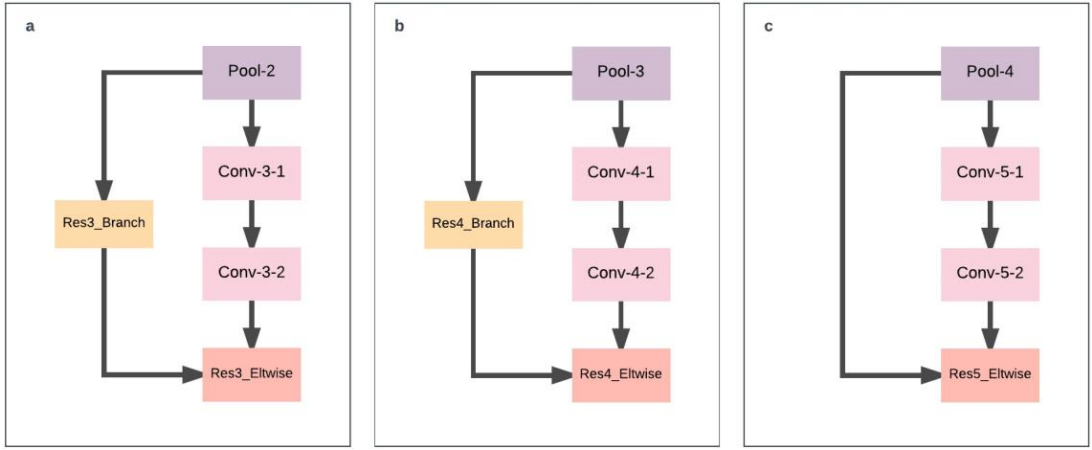

connections, the convolutional layer and the fully connected output layers. The main and auxiliary classifiers have an output layer, which has a softmax layer to compute class probability.

$$W_{main} = (W_1, \ldots, W_{13}) \quad (13)$$
$$W_{branch1} = (W_{s5}, \ldots, W_{s8}) \quad (14)$$
$$W_{branch2} = (W_{s9}, \ldots, W_{s12}) \quad (15)$$

Equation (13) [19] shows the names of weights from the main branch. As the weights in equation (13), are the weights of the 10 convolutional layers and three fully connected layers in the main branch. Furthermore, the weights of the first subsidiary branch are shown in equation (14) [19], which are the weights of one convolutional layer and three fully connected layers. Next, the weights of the second subsidiary branch are shown in equation (15) [19], which are the weights of one convolutional layer and three fully connected layers.

In the beginning, assuming the feature map created from the output layer in the main branch to be $X_{13}$, we are able to

TABLE (3)

COMPARISON OF THE TOP 1 & 5 VALIDATION CLASSIFICATION ACCURACY (%) OF OUR FINE-TUNED MODEL WITH FINE-TUNED CNDS-8 [19] NETWORK ON THE MIT PLACES365-STANDARD DATASET

| Network | Top-1 Validation | Top-5 Validation |
|---|---|---|
| CNDS-8 [19] | 54.42 | 84.71 |
| Our Model: Residual-CNDS-8 | 54.82 | 85.71 |

compute probability with the help of the softmax function from labels k =1, ..., K, as shown in equation (16) [19]. Moreover, if we assume the feature map is the first auxiliary branch $S_8$, and the feature map of the second auxiliary branch $S_{12}$ created from the output layer in both the 1st and 2nd auxiliary branches, then computing the response is granted by equations (7) and (17) [19].

$$pk = \frac{\exp(X_{13(k)})}{\sum_k \exp(X_{13(k)})} \quad (16)$$

$$psk2 = \frac{\exp(S_{12(k)})}{\sum_k \exp(S_{12(k)})} \quad (17)$$

The loss of Residual-CNDS version two can be computed utilizing the equation (8) [19] in the main branch. The loss in the main branch is calculated by utilizing likelihoods generated from the softmax function. On the other hand, the first subsidiary classifier loss is computed utilizing equation (18) [19], while the second classifier loss can be computed from equation (19) [19]. The loss generated from the subsidiary classifiers contain the weights of the subsidiary classifiers and the weights of the previous convolutional layers in the main branch.

$$L_{s1}(W_{main}, W_{branch1}) = -\sum_{k=1}^{K} yk \ln psk\,1 \quad (18)$$
$$L_{s2}(W_{main}, W_{branch2}) = -\sum_{k=1}^{K} yk \ln psk\,2 \quad (19)$$

Loss generated from the main branch and the subsidiary classifiers can be computed utilizing equation (20) [19]. Equation (20) computes weighted sum as the main branch is allotted more weight than both auxiliary classifiers. The symbol $\alpha_t$ is used to compute the value of the auxiliary classifiers as an adjustment parameter. $\alpha_t$ decays with consecutive epochs as shown by equation (11) [19], where N represents total epochs. In Residual-CNDS version two we utilize the shortcut connections derived from residual learning [22] as shown in equation (12) [22].

$$L_s(W_{main}, W_{branch1}, W_{branch2}) = L_0(W_{main}) + \alpha_t(L_{s1}(W_{main}, W_{branch1}) + L_{s1}(W_{main}, W_{branch2}) \quad (20)$$

FIG (5)

PROPOSED TEN CONVOLUTIONAL NETWORKS WITH DEEP SUPERVISION AND RESIDUAL LEARNING (RESIDUAL-CNDS). DASHED RECTANGLES SHOWS THE DEEP SUPERVISION BRANCH AND
RESIDUAL CONNECTIONS ARE THE RED

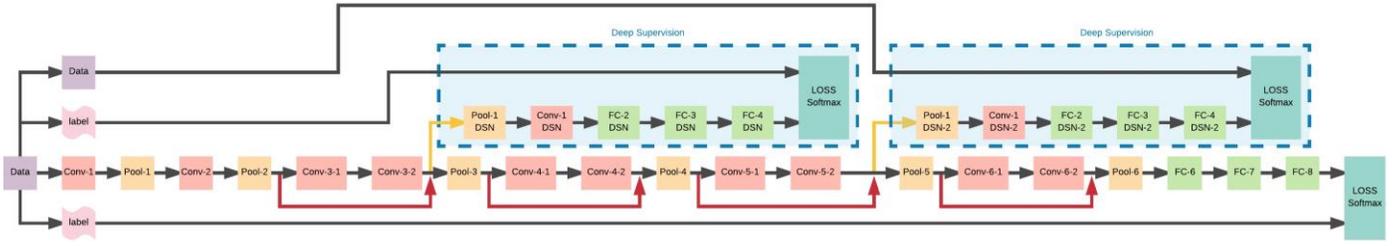

The first residual learning connection facilitates a connection between input of Conv3-1 and output from Conv3-2 using the element-wise addition, which links output from Pool-2 to the output from Conv3-2. Conv-2 and Conv3-2 have kernels of size 128 and 256 respectively. Because these two convolutional layers have unequal kernel size, we are required to form a convolutional layer of kernel size 256 between Pool-2 and element-wise addition. Figure 4 demonstrates the added convolutional layer (Res3_Branch).

Second, the second residual connection connects output of Pool-3 to output from Conv4_2, and the residual connection surpasses two convolutional layers. Hence, the residual learning connection links the input of Conv4_1 to output from Conv4-2. Furthermore, we add an adjustment convolutional layer (Res4_Branch) between Conv3-2 and Conv4-2 due to differing kernel sizes, as demonstrated in Figure 4. Thus, the element-wise addition layer can mirror the output of Conv3-2 and Conv4-2 smoothly.

We add the first auxiliary classifier next to the element-wise addition integration process after the output of Pool-3 and before Conv4-2. The third residual connection is added to connect the output from Pool-4 to Conv5-2. In the third element-wise addition integration process we do not add the adjustment convolutional layer because the kernel size of the Conv4-2 and Conv5-2 is an equal 512 for both.

We added the second and last subsidiary classifier next to the element-wise addition integration process between the output of Pool-4 and the Conv5-2. Furthermore, we attached the fourth and final residual connection after output of Pool-5 and before Conv6-2. Also, it was unnecessary to insert an adjustment convolutional layer because the kernel size of the Conv5_2 and Conv6_2 is an equal 512 for both.

### .IV. Image Dataset Description

We ran these experiments on MIT Places 205 [31], and MIT Places365-Standard [34] datasets. The CNDS, AlexNet, and GoogLeNet were all ran on the MIT places dataset, which renders it an ideal testbed for our experiments. MIT Places dataset outranks ImageNet (ILSVRC2016) [35] and SUN dataset [32] in terms of size. MIT Places has two datasets; MIT Places 205 [31], which consists of 2.5 million images from over 200 different scene categories. Image numbers in each class range from 5,000 to 15,000. All classes have 2,448,873 training images with 100 images to a category for the validation set and 200 images to a category for the test set. On the other hand, The MIT Places365-Standard [34] dataset has 1,803,460 training images while each class contains images with numbers varying ranging from 3,068 to 5,000. MIT Places365-Standard [34] dataset has 50 images/class as validation set and 900 images/class as a test set. Both MIT Places 205 [31], and MIT Places365-Standard [34] datasets are scene-centric datasets, which matches images to labels with a scene/place name. The goal of the MIT Places dataset is to assist the academic goals in the field of computer vision.

### V. Experimental Environment and Approach

Residual-CNDS version one was trained with eight convolutional layers and three residual connections in the main branch, and one convolutional layer of the subsidiary branch from scratch. In our work, we use Caffe [28], an open source deep learning framework created and developed by the Berkley Vision and Learning Center. We use a flavor of Caffe that easily adapts into the NVIDIA DIGITS deep learning GPU training system [33], which is another open source platform that allows users to build and exam their artificial neural networks for object detection and image classification with real-time visualization. For hardware we are running four NVIDIA GeForce GTX TITAN X GPUs and two Intel Xeon processors with a total of 48/24 logical/physical cores and 256 GB of hard disk space.

The images from our training, validation and testing dataset are resized to 256*256. We subtract the average pixel for each RGB color channel, the only preprocessing executed. We adjust the batch size of training phase to 256, while we adjust the batch validation size to 128. Moreover, we adjust epoch value to 50, and set learning rate to 0.01. We set the learning rate to be degraded five times during the training phase after every 10 epochs. We also adjust the decay of the learning average to half of the prior value. We cropped the images to 227*227 from random points prior to inputing them into the 1st convolutional layer. Next, the weight of all layers is set from Gaussian distribution with a 0.01 standard deviation. We adapt image reflection in the training dataset, the only image augmentation performed.

We trained the Residual-CNDS version one network on the MIT Places 205, which contains 2.4 million training elements spread accross 205 scene categories. Places 205 has 5,000-15,000 images per category. This model was validated with 100 images per category. Our model took two days and 14 hours in the training process. In the test dataset we used the epoch of the highest validation accuracy which ended up being 42 specifically. Furthermore, the subsidiary branch was used in the training phase while it was removed in the testing phase [19]. We utilized an average of the 10-crops method in testing phase, which gives an improvement over other testing methods [2]. Category labels are not available for the testing dataset and therefore the predictions that we got from the testing dataset were submitted to the MIT Places 205 server to acquire the test results as discussed in section VI.

Next, we trained AlexNet [31], GoogLeNet [31], CNDS-8[19], Residual-CNDS version one and two from scratch on MIT Places 365-Standard. We also used Caffe [28] and NVIDIA DIGITS [33] in the training phase. We utilized the same hardware, four NVIDIA GeForce GTX TITAN X GPUs and two Intel Xeon processors with 48/24 logical/physical cores and 256 GB of hard disk space.

The images in the Places 365-Standard [34] dataset for training, validation, and testing are modified to have a size of 256*256. We subtract the average pixel of each color channel of RGB color space, the only preprocessing executed. We adjust the batch size of the training phase to 256, while we set the batch size of the validation phase to 128. Moreover, we adjust the epoch count to 30, and we set learning rate to 0.01. We set the learning rate to degrade five times over the course of the training phase after every 6 epochs. Furthermore, we adjust the decay of the learning average to half that of the prior value. We cropped the images to 227*227 from random coordinates prior to passing them to the 1st convolutional layer. Next, the weight of all layers is set from Gaussian distribution with a 0.01 standard deviation. We adapted image reflection in the training dataset, the only image augmentation used.

MIT Places 365-standard [34] dataset, on which we trained AlexNet [2], GoogLeNet [7], CNDS-8 [19], Residual-CNDS version one and two, contains 1,803,460 training images and each class has images numbers ranging from 3,068 to 5,000. Places365-Standard [34] dataset contains 50 image/classes as validation set and 900 images/classes as test set. AlexNet [2], GoogLeNet [7], CNDS-8 [19], Residual-CNDS version one and two were validated with 100 images per category. AlexNet [2], GoogLeNet [7], CNDS-8 [19], Residual-CNDS version one and two took one day and 9 hours, one days and 6 hours, 22 hours, one day and 4 hours, and one day and 2 hours respectively in the training process. Class labels of the Places 365-standard are not available for the testing dataset so we are only able to report the top-1 and top-5 accuracy when performed on a validation set.

Next, we used networks CNDS-8 and Residual-CNDS version one, which are pre-trained on MIT Places 205 [31] dataset, and fine-tuned on MIT Places 365-standard [34] dataset. Epoch count is set to 20, and learning rate to 0.001. We set the learning rate to be degrade five times during the training phase after every 4 epochs. We also adjust the learning average decay to half that of the prior value. We cropped the images to 227*227 from random points prior to inputting them into to the first convolutional layer. Next, the weight of all layers is set from Gaussian distribution with a 0.001 standard deviation. We adapt image reflection in the training dataset, the only image augmentation used.

CNDS-8, Residual-CNDS version one took 14 hours, 18 hours respectively in the training process. Table (3) gives the loss and the accuracy of the networks. We report the top-1 and top-5 accuracy on the validation set for CNDS-8 and Residual-CNDS version one.

## VI. RESULTS AND DISCUSSION

In our work, we aimed to gather two strong ideas, convolutional neural networks with deep supervision [19] and residual learning [22] for training deep neural networks. We set out to see if adding residual connections to the CNDS network can boost the effectiveness of the CNDS network. We found that residual connections are parameter free connections and only add a trivial amount of complexity for the collection process, which results in it having a tiny effect on the clarity of the network. Our experiments on the MIT Places 205 and Places 365-Standard datasets back up our hypothesis that inserting the residual connections in the CNDS network will boost the accuracy of the network. We can see in table (1) that top-1 outcome of our Residual-CNDS version one, trained from scratch, exceeds the original CNDS [19] by 1.32% and 0.91% at validation and test sets respectively on Places 205 [31]. Moreover, our models top-5 results exceed the original CNDS [18] by 0.71% and 0.68% at validation and test sets respectively. Our Residual-CNDS version one model also recorded an improvement over GoogLeNet [7] and AlexNet [2] that was reported by the MIT team [31] on Places 205. The performance improvements of our network are significant when accounting for the huge obstacles that MIT Places 205 poses.

We can observe in table (2) that top-1 outcome of our Residual-CNDS version one and two, trained from scratch on MIT Places 365-standard [34] dataset, surpass the original CNDS [19] by 1.25% and 1.24% in validation set for one and two respectively on Places 365-standard [34]. Moreover, our model's top-5 results surpass the original CNDS [18] by 1.08% and 1.25% at validation set for one and two respectively. Furthermore, our models (Residual-CNDS eight and ten) exceed the AlexNet [2] and GoogLeNet [7] in both top-1 and top-5.

Finally, our fine-tuned (Residual-CNDS eight layers), which was pre-trained on MIT Places 205 and applied on MIT Places 365-standard [34] dataset, exceeds CNDS [19] by 0.4% and 1% in top-1 and top-5 respectively. Table (3) shows the results of our fine-tuned model.

We can say that our proposed Residual-CNDS version one and two (eight and ten layers) have a better performance than currently top performing neural networks. Furthermore, we applied our model on very large scale scene image datasets, which imposes a great challenge for any neural network, nevertheless, our proposed models still give good performance. Given all of the above, we are confident that our models gather the best of both CNDS and residual learning practices, making it easy to converge and override the overfitting and degradation problems.

## VII. Conclusion and Future Work

We've hereby introduced two flavors of the Residual-CNDS network: eight and ten layers, in which we add residual learning using shortcut connections. In our paper, we implemented our models on the massive image datasets MIT Places 205 [31] and Places 365-Standard [34], which shows that the proposed networks exceed recent high-grade networks in both top-1 and top-5 accuracy.

Future research will focus on the impact of residual learning on other widespread networks including AlexNet and VGG.